\documentclass[journal,twoside,web,letter]{ieeecolor}
\usepackage{generic}
\usepackage{cite}
\usepackage{amsmath,amssymb,amsfonts}
\usepackage{algorithm,algorithmic}
\usepackage{textcomp}
\usepackage[skins]{tcolorbox}
\usepackage{CJK}
\def\BibTeX{{\rm B\kern-.05em{\sc i\kern-.025em b}\kern-.08em
    T\kern-.1667em\lower.7ex\hbox{E}\kern-.125emX}}
\usepackage{color, colortbl}
\usepackage{lipsum}
\usepackage{xspace}
\usepackage{dsfont}
\usepackage{amssymb}
\usepackage{multirow}
\usepackage{stmaryrd}
\usepackage{booktabs}
\usepackage{graphicx}
\usepackage{nicefrac}
\usepackage{relsize}
\usepackage{mathabx}
\usepackage{array}
\usepackage{calc}
\usepackage{ifthen}
\usepackage{makecell}
\usepackage{fancyhdr}
\usepackage[utf8]{inputenc}
\usepackage{pifont}

\newcommand{\zalm}{\textcolor{black}ZALM3\xspace}

\begin{document}

\title{ZALM3: Zero-Shot Enhancement of Vision-Language Alignment via In-Context Information in Multi-Turn Multimodal Medical Dialogue}

\author{
Zhangpu Li$^{1\dagger}$, Changhong Zou$^{1\dagger}$, Suxue Ma$^{2\dagger}$, Zhicheng Yang$^{3\ddagger}$, Chen Du$^3$, Youbao Tang$^3$, Zhenjie Cao$^4$, Ning Zhang$^3$, Jui-Hsin Lai$^3$, Ruei-Sung Lin$^3$, Yuan Ni$^4$, Xingzhi Sun$^4$, Jing Xiao$^4$, Jieke Hou$^5$, Kai Zhang$^{1\ddagger}$, and Mei Han$^3$
\thanks{$^\dagger$This work was done during Zhangpu Li, Changhong Zou, and Suxue Ma's internship at Ping An Technology, Shenzhen, China.}
\thanks{$^\ddagger$Corresponding authors: \texttt{zcyangpingan@gmail.com}; \texttt{zhangkai@sz.tsinghua.edu.cn}}
\thanks{$^1$Z.L., C.Z., and K.Z. are with Tsinghua Shenzhen International Graduate School, China.}
\thanks{$^2$S.M. is with Tianjin University, China.}
\thanks{$^3$Z.Y., C.D., Y.T., N.Z., J-H.L., R-S.L., and M.H. are with PAII Inc., USA.}
\thanks{$^4$Z.C., Y.N., X.S., and J.X. are with Ping An Technology, China.}
\thanks{$^5$J.H. is with Ping An Healthcare and Technology Company Limited, China.}
}

\maketitle

\begin{abstract}
The rocketing prosperity of large language models (LLMs) in recent years has boosted the prevalence of vision-language models (VLMs) in the medical sector. In our online medical consultation scenario, a doctor responds to the texts and images provided by a patient in multiple rounds to diagnose her/his health condition, forming a multi-turn multimodal medical dialogue format. Unlike high-quality images captured by professional equipment in traditional medical visual question answering (Med-VQA), the images in our case are taken by patients' mobile phones. These images have poor quality control, with issues such as excessive background elements and the lesion area being significantly off-center, leading to degradation of vision-language alignment in the model training phase. In this paper, we propose ZALM3, a Zero-shot strategy to improve vision-language ALignment in Multi-turn Multimodal Medical dialogue. Since we observe that the preceding text conversations before an image can infer the regions of interest (RoIs) in the image, ZALM3 employs an LLM to summarize the keywords from the preceding context and a visual grounding model to extract the RoIs. The updated images eliminate unnecessary background noise and provide more effective vision-language alignment. To better evaluate our proposed method, we design a new subjective assessment metric for multi-turn unimodal/multimodal medical dialogue to provide a fine-grained performance comparison. Our experiments across three different clinical departments remarkably demonstrate the efficacy of ZALM3 with statistical significance.
\end{abstract}

\begin{IEEEkeywords}
multi-turn multimodal dialogue, zero-shot vision-language alignment, subjective assessment
\end{IEEEkeywords}

\section{Introduction}
\label{sec:intro}
Recent years have witnessed the tremendous success of large language models (LLMs) \cite{achiam2023gpt,touvron2023llama,jiang2023mistral,penedo2023refinedweb,workshop2022bloom}. They have even shown their massive potential in various professional application domains when equipped with specialized knowledge, such as legal \cite{sun2023short}, financial \cite{li2023large}, and medical sectors \cite{labrak2024biomistral,singhal2023large,thirunavukarasu2023large,singhal2023towards}. 
A challenging format in these applications is multi-turn dialogue, where an LLM model-based assistant interacts with a user in multiple rounds to finally fulfill the user's request. Multi-turn dialogue requires the model not only to understand the user's current input but also to provide coherent responses by considering the dialogue history. This has raised higher requirements for the model's contextual understanding and generation capabilities. In the medical field, with the worldwide lockdown and quarantine caused by COVID-19, online doctor consultation applications have rapidly developed. Recently, some cutting-edge studies have focused on real multi-turn medical dialogue \cite{yunxiang2023chatdoctor,han2023medalpaca,yang2024zhongjing,chen2023huatuogpt,chen2023bianque}, however, their conversations involve only text and lack the ability to handle multiple modalities, such as images sent by users. 

Meanwhile, scholars have intensively investigated the integration of visual modality with LLMs to achieve vision-language models (VLMs) by leveraging Transformer technology and other techniques \cite{yin2023survey,liu2024visual,li2023blip,zhu2023minigpt,alayrac2022flamingo}. 
As the medical field is a significant branch for VLM-based applications \cite{tu2024towards,saab2024capabilities,moor2023med}, numerous methods have recently emerged to effectively learn visual and textual information (e.g. radiology reports) using VLMs to enhance the performance of medical imaging analysis \cite{guo2023multiple,lu2024visual,mahapatra2023class,huang2024adapting}, medical report generation \cite{wang2023fine,liu2023observation,wu2023improving}, and medical visual question answering (Med-VQA) \cite{li2023masked,he2024pefomed}. 
Nevertheless, the dominant settings were single-turn \cite{zhang2024sam,jin2024promptmrg} or multi-turn stitched by independent single-turn pairs of visual questions and textual answers \cite{lau2018dataset,he2020pathvqa}. 
Different from conventional multi-turn textual medical dialogue or Med-VQA settings, our scenario involves multi-turn medical consultations with text and image modalities together, unveiling a multimodal research branch that has not been well addressed.

In our case, patients describe their conditions by sending texts and images, while doctors ultimately provide diagnoses, medication recommendations, or referrals via multiple rounds of interaction with patients. Distinct from those professional medical images \cite{guo2023multiple,lu2024visual,mahapatra2023class,huang2024adapting,wang2023fine,liu2023observation,wu2023improving,li2023masked,he2024pefomed,zhang2024sam,jin2024promptmrg,lau2018dataset,he2020pathvqa}, all images in our setting are provided by patients with their phones. As a result, those images may contain significant noises, such as excessive background elements and non-centered regions of interest (RoIs). For a real doctor, such a poor-quality image is not a hurdle as long as a lesion in the image is discernible. However, noisy images significantly degrade the vision-language alignment for training a VLM, detrimentally affecting the performance of the model's response.

To enhance vision-language alignment performance, recent approaches have proposed various adaptations based on CLIP (Contrastive Language-Image Pre-training) \cite{wang2022medclip,you2023cxr,chen2024mammo,zhang2023biomedclip}. 
Some other representative methods include projecting visual features \cite{liu2024visual,li2023blip,bai2023qwen}, employing multiple vision encoders \cite{tong2024eyes}, or decoupling the language and vision training process \cite{jian2024bootstrapping}.
Unlike those approaches that need to alter the structure of VLMs, we employ a \emph{zero-shot visual grounding} method to update images by extracting their key regions. Visual grounding is mapping a natural language query to the specific visual region that the query refers to \cite{yang2021visual}. 
It always requires an external word or phrase input as the textual query.
Since a multi-turn dialogue maintains semantic coherence, when a patient sends an image related to her/his medical condition, the descriptions of this image have always been mentioned in the prior dialogue turns. For instance, a patient in the dermatology department may describe her/his symptoms and affected areas in the text before sending an image. This preceding context acts as a \emph{hidden gem}, which can be leveraged as a textual query for visual grounding inferring the key regions of the image. With this procedure, we do not need 1) human annotation labor to crop the image (thanks to the visual grounding approach) or 2) an external query for visual grounding (thanks to the preceding context of the image). Furthermore, this paradigm simulates the real diagnostic routine where a doctor gathers preliminary information through text communications before viewing the image. Consequently, even if the image has high background noise, the doctor can still ignore the noise and focus on the relevant areas based on prior knowledge.

In this paper, we present \zalm (\textbf{Z}ero-shot enhancement of vision-language \textbf{AL}ignment in \textbf{M}ulti-turn \textbf{M}ultimodal \textbf{M}edical dialogue). Our key contributions are as follows.
\begin{itemize}
	\item We propose an LLM-powered and visual grounding-based approach to enhance the vision-language alignment in the unprecedented multi-turn multimodal medical dialogue scenario. It leverages the in-context information before the image's appearance without introducing extra annotation or training efforts. This approach is a zero-shot and compatible with various medical VLMs.
	\item We reformulate the subjective assessment strategy for the multi-turn unimodal/multimodal medical dialogues, and design new evaluation metrics for the subjective assessment to achieve more quantitative results over the previous simplistic win-loss-tie evaluation criteria.
	\item We develop two versions of medical VLMs and conduct extensive experiments on our multi-turn multimodal medical dialogue datasets. The results demonstrate the efficacy of \zalm with statistical significance across different clinical departments.
\end{itemize}

\section{Related Work}
\label{sec:related}

\textbf{Med-VQA and Medical Dialogue.} VQA is an interdisciplinary task that requires the capacities of both computer vision (CV) and natural language processing (NLP). As a significant branch of VQA, Med-VQA has garnered widespread attention from scholars. Many public datasets were released to provide specialized data support for the Med-VQA domain \cite{liu2021slake,huang2022ovqa,lau2018dataset,he2020pathvqa}. However, these dialogue datasets are either single-turn (e.g. medical image captioning or medical report generation) or multi-turn by simply concatenating single-turn question-answer pairs without direct contextual relevance. Furthermore, these datasets are constructed in an image-centered format, rather than being built based on the dialogue of real patient-doctor interactions.
Since medical LLMs have been investigated in recent years for various NLP-based tasks \cite{gu2021domain,venigalla2022pubmed}, numerous studies have explored the capacity of the LLMs for medical dialogue in English or other languages such as Chinese \cite{singhal2023large,singhal2023towards,yang2024zhongjing,chen2023huatuogpt,wang2023huatuo,zhang2023huatuogpt,chen2023bianque}. Moreover, researchers leveraged the capability of multimodal LLMs to tackle the image-based tasks such as case diagnosis and radiology or pathological image classification and recognition \cite{guo2023multiple,lu2024visual,mahapatra2023class,huang2024adapting,li2023masked,he2024pefomed}. In those studies, medical images were obtained from professional equipment with fixed fields of view and angles, where a high image quality was ensured. However in our scenario, images are captured by patients' mobile phones, resulting in significant noise that adversely affects vision-language alignment at the phase of model training.

\begin{figure*}[t]
    \centering
    \includegraphics[width=\linewidth]{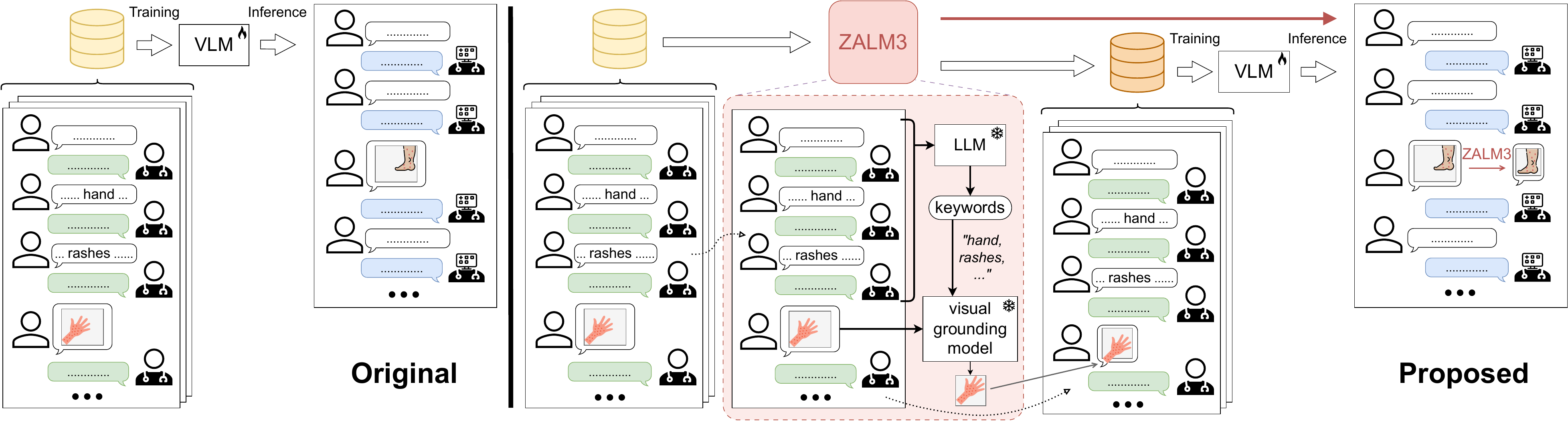}
    \vspace{-3ex}
    \caption{\textbf{Framework of the proposed \zalm.} In this example, compared with the original VLM training strategy, our \zalm enhances vision-language alignment by extracting the relevant region from images sent by patients in a zero-shot manner, which includes an LLM and a visual grounding model (frozen in the red block). The LLM extracts keywords from the preceding context before the image. The keywords are fed into the visual grounding model to crop the image.} The cropped image updates the original one, and the updated database is used for VLM training. \zalm is also used during the inference process (long red arrow). 
    \vspace{-2ex}
\label{fig:framework}
\end{figure*}

\textbf{Visual-Language Alignment.} Aligning the features of image and text is essential for the VLM research domain \cite{du2022survey}. The pioneering CLIP leveraged enormous image-text pairs to validate effective vision-language alignment \cite{radford2021learning}. By encoding medical domain knowledge, various contributions adapted from CLIP aimed to solve general \cite{zhang2023biomedclip} or specific problems in medical image analysis, such as chest X-rays \cite{wang2022medclip,you2023cxr} and mammograms \cite{chen2024mammo}. A recently advanced paradigm designed projection layers for visual features, such as Bootstrapping Language-Image Pre-training v2 (BLIP-2) \cite{li2023blip}, Large Language and Vision Assistant (LLaVA) \cite{liu2024visual}, and Qwen-VL \cite{bai2023qwen}, leading to further specific modifications on them for medical applications \cite{chen2023medblip,li2024llava,liu2023qilin}. Other methods included introducing multiple vision encoders to extract diverse visual features \cite{tong2024eyes}, or decoupling the training process of language and vision components \cite{jian2024bootstrapping}. Different from those methods that focus on the improvement at the model level, our proposed approach mines the in-context information for image refinement to achieve better vision-language alignment.

\textbf{Visual Grounding.} Another mindset to facilitate vision-language alignment is to emphasize the RoIs \cite{reich2024uncovering}. Some VLMs have possessed the region-level or pixel-level grounding ability to output the RoIs in bounding box or segmentation mask formats \cite{bai2023qwen,rasheed2023glamm,zhang2023gpt4roi,peng2023kosmos}. Meanwhile, outstanding visual foundation models like Grounding DINO (GDINO) \cite{liu2023grounding} and Segment Anything Model (SAM) \cite{kirillov2023segment} have provided zero-shot grounding capacities. In the medical sector, aside from initially testing a grounding model's zero-shot performance on medical image datasets \cite{he2023accuracy,huang2024segment}, researchers proposed an enhanced framework for chest X-ray segmentation by jointly using GDINO and SAM \cite{ramesh2023lung}. Authors in \cite{wang2023sam} proposed a SAM-assisted approach for enhancing medical image annotation. However, those methods required externally well-defined prompts to trigger the grounding models, likely involving non-negligible human effort. In contrast, our grounding prompts in \zalm are entirely derived from the preceding context in the multi-turn dialogue without human intervention.

\section{Materials and Methods}
\label{sec:method}

\subsection{Overview of the Proposed Method}
Fig.~\ref{fig:framework} illustrates our proposed method. In an original VLM training strategy, the multi-turn multimodal conversation data between doctors and patients is directly fed into the VLM model for training. 
Once a VLM model is trained, multi-turn conversation inference is performed. In our proposed framework, the key module, \zalm, eliminates the irrelevant regions from the images in the original database, while preserving their original positions in the conversations. This process does not need any training procedure. The updated database is then used for VLM training, where the vision-language alignment is improved by extracting the RoIs from the original images \cite{yang2023set,cai2023making,wan2024contrastive}. Once the enhanced VLM training is finished, the \zalm module is also used during inference, ensuring that the data processing during inference is consistent with that during training.

\subsection{Description of Database}
\label{sec:database}
Our data comes from a leading Chinese online medical consultation platform, covering over 10 major clinical departments. Nationwide patients interact with doctors through multiple turns of online conversations to achieve diagnoses, prescriptions, referrals, and other medical purposes. In addition to pure textual conversations, patients also send images in the dialogue when necessary to provide further explanations or evidence to the doctors. This platform has over 3.2M consultation \textit{sessions} per month, including more than 1.5M images on average. While the quality of text content is acceptable, the quality of images uploaded by patients is significantly lower. Since those images are taken with the patients' mobile phones, many of them are adversely affected by factors such as lighting, angles, resolution, etc. For instance, in the department of dermatology, it is challenging for patients to independently capture certain body parts, resulting in the RoI not being centered in the image. In such cases, a real doctor can identify the body part in the image, but it is difficult for an AI model. 

\begin{figure}[t]
    \centering
    \vspace{-1ex}
    \includegraphics[width=\linewidth]{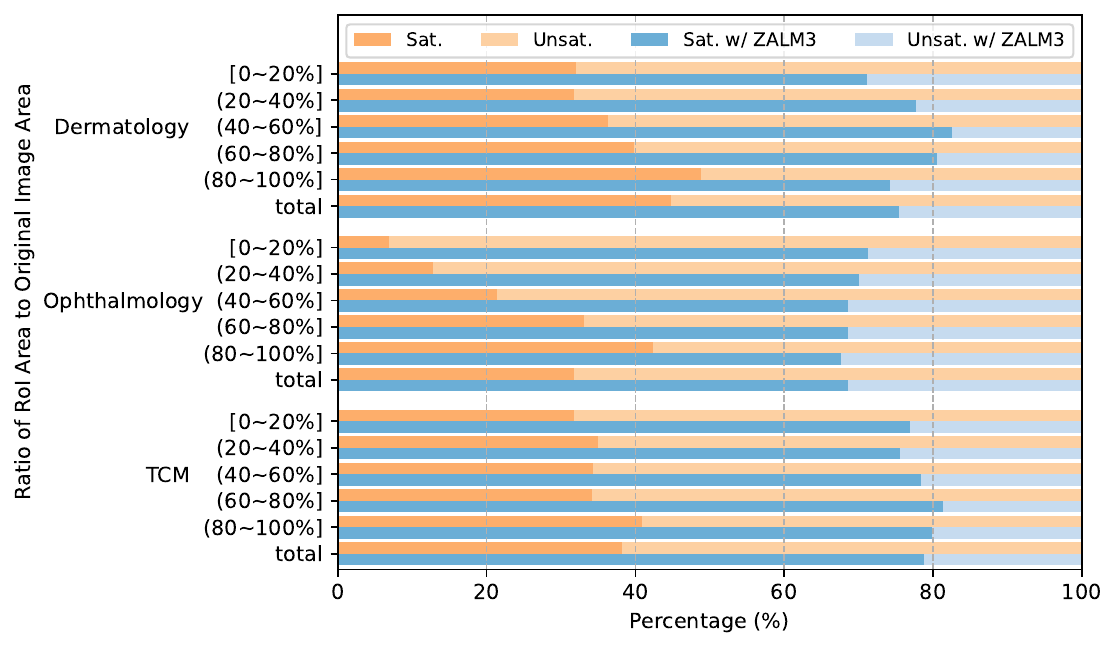}
    \vspace{-4.7ex}
    \caption{Percentages of satisfactory/unsatisfactory image quality based on five area ratio segments of RoIs to original images across different clinical departments without and with \zalm .} 
    \vspace{-3.8ex}
\label{fig:barchart}
\end{figure}

In several departments where images constitute a significant portion of the data, including dermatology, ophthalmology, and traditional Chinese medicine (TCM), we recruited 12 volunteers to evaluate the quality of approximately 60k images. They assessed issues such as excessive background information, blurriness, and misalignment of the RoI from the center of the image. Fig.~\ref{fig:barchart} presents the assessment results based on the 5 area ratio segments of RoIs to original images across the three departments. As we can see, the rates of poor-quality images for all three departments are significantly reduced by our proposed \zalm under all ratios.

\subsection{Extraction of Keywords}
\label{sec:keyword}
In our conversational scenario, patients initially describe their symptoms in text and doctors respond accordingly. When the patient needs to upload an image for further diagnosis, the description of this image is actually reflected in the preceding context. For example, when a patient initially describes that ``my leg got bitten by a mosquito and it was swollen'', this has already provided some description of the upcoming image the patient will send. Therefore, an extra manual annotation for the image's RoIs is not necessary. While we leverage the preceding context information to generate the description of the image, these conversation histories contain a lot of noise and unrelated content. To achieve more appropriate inputs for a visual ground model, 
we exploit an LLM to extract keywords from the conversation history preceding the image. To balance the precision and processing time of keyword extraction, the previous up to three turns of patient-doctor conversations in the text before an image are utilized, as we observe that the relevance between the earlier conversations and the image is not significant. Once the keywords are extracted, they are then fed into the visual grounding model.

\subsection{Visual Grounding}
\label{sec:gdino}
A visual grounding task aims to identify the pertinent object or region in an image by a natural language query \cite{ichinose2023visual}. We leverage GDINO \cite{liu2023grounding} as a zero-shot object detector to tackle this task. GDINO combines the Transformer-based object detection algorithm DETR with Improved DeNoising (DINO) \cite{zhang2022dino} and the grounding vision-language pre-training model Grounded Language-Image Pre-training (GLIP) \cite{li2022grounded}, equipped with the capacities of understanding natural language queries and detecting the relevant objects in an image. Specifically, GDINO first extracts vanilla image and text features, enhancing them via a feature enhancer module to obtain cross-modality features. It then selects queries from the image features by a language-guided query selection module, feeding them into a cross-modality decoder to predict object boxes with related phrases.

Once the keywords are given from Sec.~\ref{sec:keyword}, the visual grounding model outputs the object detection results with bounding boxes and the corresponding keywords.  Those outlined regions are kept as our image refinement result. For multiple bounding box outputs (e.g. multiple keywords are found, or multiple objects for one keyword are found), we take the union of coordinates of these bounding boxes to accommodate all their regions. If any keyword or any object bounding box is not activated in this image, the image will be kept as the same as the original one. Hence, our visual grounding procedure is conservative to ensure that as much information as possible from the image is retained and not overlooked. Once visual grounding is complete, the updated image replaces the original in the same conversation spot.

\subsection{Medical VLM}
\label{sec:medvlm}
Although many Chinese medical LLMs have emerged \cite{chen2023bianque,zhang2023huatuogpt,yang2024zhongjing,xiong2023doctorglm,wang2023huatuo}, there is still a gap in Chinese medical VLMs. To address this, we have developed two versions of medical VLM frameworks along the timeline of LLM development. The initial framework replaces the English LLM in BLIP-2 \cite{li2023blip} with the Baichuan2-13B-Chat \cite{baichuan2023baichuan2} model. The current framework is based on the Qwen-VL architecture \cite{bai2023qwen}, upgrading the LLM component from Qwen-7B-Chat to Qwen-14B-Chat \cite{qwen}. 
The details of training and inference strategies are described in Sec.~\ref{sec:imple}.
Note that 
our proposed method is compatible with other medical LLMs, such as the recently released HuatuoGPT-II series \cite{chen2023huatuogpt}, which features a larger backbone (34B) and requires more computational resources.

\section{Design of Subjective Assessment}

\subsection{Difficulty of Objective Assessment}
\label{sec:objective}
In single-turn question-answering tasks, such as an image captioning task, a model's generated response can be directly compared for similarity with the ground truth description, and further a numeric accuracy result can be achieved. However, in multi-turn medical dialogue tasks, the ground truth for each dialogue turn is difficult to define precisely. For instance, before providing a diagnosis, a doctor may need to ask several relevant questions to the patient, and the order of these questions is not strictly required, as long as enough information is gathered from the patient. Hence, it is inappropriate to rigidly compare the similarity between each turn of the model-generated response and the doctor's response. Another seemingly intuitive approach is to compare the correctness of the entire dialogue's diagnosis result. However, in our actual scenario, disease diagnosis or medication advice must not be independently provided by an AI model, because it would raise concerns about medical accidents and liability issues \cite{cestonaro2023defining,park2023ai}. When an AI model obtains enough information from multi-turn dialogues with patients, regardless of whether its diagnosis is correct or not, a real doctor will intervene to make the decision. Therefore, based on the aforementioned aspects, objective evaluation methods are not quite appropriate in the context of our multi-turn medical dialogue.

\subsection{Improvement of Subjective Assessment}
The International Telecommunication Union Radiocommunication Sector (ITU-R) describes methodologies for subjective assessment (ITU-R BT.500), which were originally used for assessing the quality of television images \cite{sector2023recommendation}. Among those methods, \emph{double-stimulus} methods were endorsed to compare the outputs of the two \textit{test conditions} simultaneously. This strategy has been utilized in various quality assessment tasks \cite{chow2016review,ogino2022simultaneous}.
For medical LLM quality assessment, a simplified form (e.g. a win-loss-tie percentage) of the stimulus-comparison methods in ITU-R BT.500 \cite{sector2023recommendation} has been adopted in recent Chinese medical LLMs \cite{chen2023bianque,zhang2023huatuogpt,yang2024zhongjing,chen2023huatuogpt}. 
To enhance the current performance assessment scheme in the medical VLM/LLM domain, we leverage the philosophy of double-stimulus continuous quality-scale (DSCQS) \cite{sector2023recommendation}, utilize the mechanism of differential mean opinion scores (DMOS) \cite{streijl2016mean,rassool2017vmaf}, and integrate the necessary medical domain knowledge to provide more fine-grained and robust evaluation results.

\begin{table}[t]
	\centering
	\caption{Five-grade rating table and corresponding descriptions for multi-turn unimodal/multimodal medical dialogue. 
        \vspace{-1ex}
         }
	\begin{tabular}{cl}
		\toprule
		Score & \multicolumn{1}{c}{Description} \\ \midrule
		4 & \makecell[l]{
			- The model's response was appropriate to the consultation \\ scenario at the time and aligned with medical domain \\ knowledge.
		} \\ \midrule
		3 & \makecell[l]{
			- The model's question was irrelevant to the illness; \\
			- The model should have responded to the patient's messages, \\ but it did not.
		} \\ \midrule
		2 & \makecell[l]{
			- Repetitive questioning in the model's response. For example, \\ the patient has already sent the image, but the response still \\ requested the image; \\
			- The model's answer was irrelevant to the patient's question.
		} \\ \midrule
		1 & \makecell[l]{
			- The model provided judgments with excessive confidence. \\
			- The model's response had medical terminology errors.
		}  \\ \midrule
		0 & \makecell[l]{
			- In the response attitude, there were emotions such as \\ disputes, insults, provocations, and instigation; \\
			- The confusion of communication targets was not promptly \\ corrected. For example, mistakenly treating a child as an adult \\ during a consultation; \\
			- Diseases not within the department's specialty were still \\ being treated. 
		} \\
		\bottomrule
	\end{tabular}
	\label{tab:rating}
        \vspace{-3.5ex}
\end{table}

\subsection{Proposed Evaluation Metrics}
For each doctor's response, we have two model-generated results: the original model, and the model with \zalm. Specialists rate the content generated by these two models according to a five-grade scale (0$\sim$4 points) evaluation table, which is designed via our extensive communications with doctors. By adequately integrating the doctors' necessary and professional concerns, our five-grade criteria for the multi-turn multimodal medical dialogue settings are detailed in Table~\ref{tab:rating}.

For the model with \zalm, $r_{nsm}$ represents the score given by the $n$-th evaluator for the $m$-th response in the $s$-th entire multi-turn consultation session, where $r_{nsm} \in \{0, 1, 2, 3, 4\}$, $n \in \{1, ..., N\}$, $s \in \{1, ..., S\}$, and $m \in \{1, ..., M_s\}$. Correspondingly, for the original model, the score is denoted as $r^{\text{ref}}_{nsm}$. We then calculate the difference between the two scores:
\begin{equation}
d_{nsm} = r_{nsm} - r^{\text{ref}}_{nsm}.
\end{equation}
Next, we design two types of DMOS, which are session-level ($\mathcal{D}^{\text{sess}}$) and image-level ($\mathcal{D}^{\text{img}}$), respectively: 
\begin{equation}
\mathcal{D}^{\text{sess}} = \left\{\frac{1}{N}\frac{1}{M_s}\sum_{n=1}^{N}\sum_{m=1}^{M_s}d_{nsm} ~|~ s \in \{1, ..., S\}\right\},
\end{equation}
\begin{equation}
\mathcal{D}^{\text{img}} = \left\{\frac{1}{N}\sum_{n=1}^{N}d_{ni} ~|~ i \in \{1, ..., I\}\right\},
\end{equation}
where $d_{ni}$ denotes the score difference for the $i$-th image sent by the patient; $I = \sum_{s=1}^{S}I_s$; $I_s$ represents the number of images in the $s$-th entire multi-turn consultation session.

\section{Experiment Results}
\label{sec:result}

\subsection{Implementation Details}
\label{sec:imple}
Although some Chinese multi-turn textual medical dialogue datasets are available \cite{zeng2020meddialog,xu2023medical,zhang2023huatuogpt,li2021semi}, to the best of our knowledge, there are no publicly accessible Chinese multi-turn multimodal medical dialogue datasets. Therefore, we comprehensively conduct experiments on our in-house datasets. This study is reviewed and approved by the Ethics Committee of Qingdao Ping An Kangjian Internet Hospital (IRB number: LLSC2024A01). 

Our data for the experiments includes the departments of dermatology, ophthalmology, and TCM, because these departments possess substantial amounts of image data as mentioned in Sec.~\ref{sec:database}. Moreover, the rate of poor image quality is high as shown in Fig.~\ref{fig:barchart} (orange bars). For example, \textit{tongue diagnosis} is an indispensable routine examination in TCM \cite{anastasi2009understanding}. It plays a crucial role in the diagnosis and treatment of various diseases by combining tongue examination with clinical assessments. While a doctor requires a close-up of a patient's tongue, the patient always takes a photo that includes her/his face, making the tongue region much smaller than what the doctor expects. A visualized example in Sec.~\ref{sec:frame} highlights that such vision-language misalignment downgrades the model's response.

Our servers are equipped with NVIDIA V100 16G GPUs, and have Ubuntu 18.04, Python 3.9.0, and CUDA 11.8. Due to computational limitations, our training strategy is divided into two steps. First, we fine-tune the LLM component of the VLM with LoRA (Low-Rank Adaptation) \cite{hu2021lora} using multi-turn textual dialogue data. In each department, we extract 100k entire consultation sessions that purely have text conversations (i.e. unimodal textual data), and split them by a ratio of 9:1 into training and validation sets. As a result, a dedicated LLM is obtained for each department. Second, we train the projection layer in the VLM using multimodal data. This approach ensures that we can retain enough image tokens without causing an out-of-memory (OOM) error. In each department, we extract 10k entire consultation sessions that have at least one image sent by patients. We split 9k:900:100 for training, validation, and test sets, respectively. For the VLM training, we adopt OneCycleLR \cite{smith2019super} with the max learning rate $5\times10^{-5}$ for both the models of BLIP-2 with Baichuan2-13B-Chat and Qwen-VL with Qwen-14B-Chat. The remaining parameters are set by default. In the model inference process, we concatenate the preceding text and image data from the patient and real doctor's conversations as the history input to generate the response to the user's current message. We recruit 4 specialists for subjective assessment of the model.

In \zalm, we use the Baichuan2-13B-Chat model as the LLM in our experiments. Since this LLM is independent of the VLM, it can also adopt other LLMs. Due to the page limit, the prompt is provided in our supplementary materials. We limit the number of keywords to 5. As mentioned in Sec.~\ref{sec:gdino}, GDINO is utilized as the visual grounding model. 
Although GDINO has shown the performance of their Tiny, Base, and Large models \cite{liu2023grounding}, the available checkpoints are for the Tiny (GDINO-T) and Base (GDINO-B) models only. 
We use the GDINO-B to achieve better performance. The illustrated comparisons of GDINO-T and GDINO-B models for visual grounding are presented in our supplementary materials. The bounding box threshold is set as 0.35 and the text threshold is set as 0.25. The remaining parameters of GDINO are default. 

\begin{table*}[t]  
    \centering
    \caption{Performance comparisons of LLMs with and without \zalm on various clinical departments using our multi-turn multimodal medical dialogue datasets. 
    \vspace{-1ex}
    }
    \label{tab:dmos}
    \begin{tabular}{ccllc}
    \toprule
    \multirow{2}{*}{Department} & \multirow{2}{*}{VLM} & \multicolumn{3}{c}{Averaged DMOS$\uparrow$ ($p$-value)} \\ \cmidrule{3-5}
    ~ & ~ & \multicolumn{1}{c}{Session-Level ($\overline{\mathcal{D}^{\text{sess}}}$)} & \multicolumn{1}{c}{Image-Level ($\overline{\mathcal{D}^{\text{img}}}$)} & Cropped Image-Level ($\overline{\mathcal{D}^{\text{img}'}}$)\\ \midrule
        \multirow{2}{*}{Dermatology} & BLIP-2\cite{li2023blip} + Baichuan2-13B-Chat\cite{baichuan2023baichuan2} & 0.434 (1.12$\times$10$^{-11}$) & 0.693 (3.73$\times$10$^{-10}$) & 0.756 (4.69$\times$10$^{-3}$) \\
        ~ & Qwen-VL\cite{bai2023qwen} + Qwen-14B-Chat\cite{qwen} & 0.482 (1.39$\times$10$^{-17}$) & 0.764 (1.65$\times$10$^{-16}$) & 0.804 (1.19$\times$10$^{-4}$) \\  \midrule
        Ophthalmology & Qwen-VL\cite{bai2023qwen} + Qwen-14B-Chat\cite{qwen} & 0.362 (1.35$\times$10$^{-9}$) & 0.680 (2.03$\times$10$^{-8}$) & 0.802 (1.35$\times$10$^{-6}$) \\  \midrule
        TCM & Qwen-VL\cite{bai2023qwen} + Qwen-14B-Chat\cite{qwen} & 0.230 (1.18$\times$10$^{-7}$) & 0.309 (7.35$\times$10$^{-3}$) & 0.523 (2.14$\times$10$^{-2}$) \\
        \bottomrule
    \end{tabular}    
    \vspace{-0.7ex}
\end{table*}

\begin{figure*}[t]
    \centering
    \includegraphics[width=\linewidth]{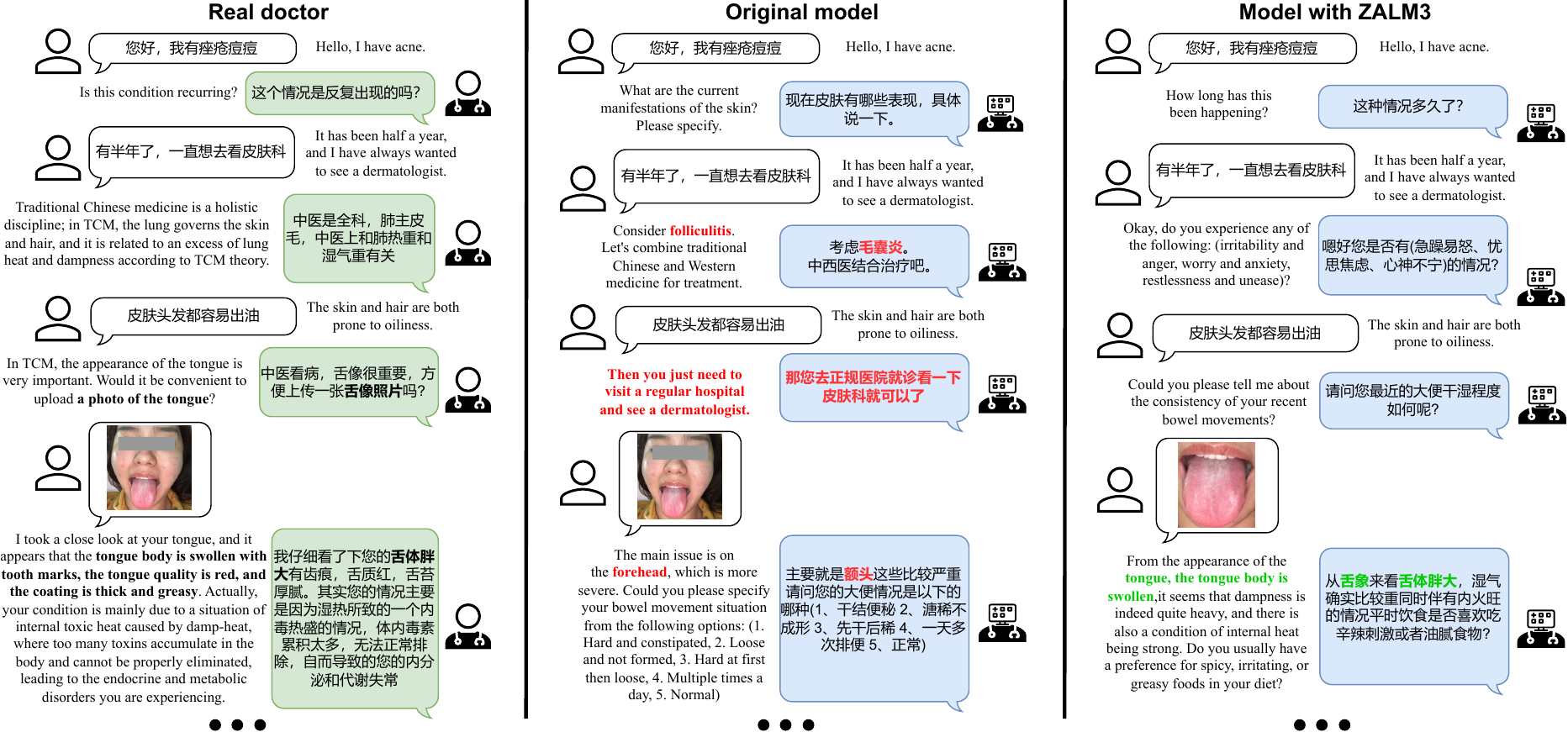}
    \vspace{-3.2ex}
    \caption{Visualized inference example from the test set from the department of TCM in our multi-turn multimodal dialogue dataset. \emph{Left}: conversation between the patient and the real doctor's response as a reference. \emph{Middle}: Original model response without \zalm. \emph{Right}: Model with \zalm. \textcolor{green}{Green} text indicates a satisfactory response (score 3$\sim$4 in Table~\ref{tab:rating}) in the key location compared with the patient-doctor reference, while \textcolor{red}{red} text denotes an unsatisfactory response (score 0$\sim$2 in Table~\ref{tab:rating}). The gray patches on the patient's eyes are manually added to protect her privacy. 
    \vspace{-2.8ex}
    }
\label{fig:compare_training}
\end{figure*}

\subsection{Results}

\subsubsection{Multi-Turn Multimodal Medical Dialogue}
\label{sec:multivqa}

We design three subjective assessment criteria to evaluate our proposed approach. We calculate the averaged values $\overline{\mathcal{D}^{\text{sess}}}$ and $\overline{\mathcal{D}^{\text{img}}}$ to represent the performance of \zalm at the session level and image level, respectively. In addition, we compute one more averaged value $\overline{\mathcal{D}^{\text{img}'}}$, which represents images where the new image area is less than 70\% of the original image area after applying \zalm, indicating that these images have been obviously cropped by \zalm. Table~\ref{tab:dmos} shows the results of these three assessments on our multi-turn multimodal medical dialogue datasets from different clinical departments. We can see that all averaged DMOS values are greater than 0, and the $p$-values are remarkably less than 0.05. This indicates that our method is drastically effective across different departments. 
In particular, the session-level improvement is most robust according to the highly significant $p$-values. Furthermore, we observed that the results at the cropped image-level column are the highest, confirming that these images cropped by \zalm enhance vision-language alignment, which benefits the model's responses to other images and further improves the model's performance on the entire session.

In Sec.~\ref{sec:medvlm}, we mentioned the two versions of our medical VLM. Besides applying our current framework (Qwen-VL\cite{bai2023qwen} + Qwen-14B-Chat\cite{qwen}) to three departments, it is important to note that the initial framework (BLIP-2\cite{li2023blip} + Baichuan2-13B-Chat\cite{baichuan2023baichuan2}) was only applied to the department of dermatology. One reason is that the first dataset we obtained was from the dermatology department, and the data from the two other departments were not available to us at the same time. Another reason is that we found the performance of the current framework was significantly better than that of the initial one. The results of this observation will be described in Sec.\ref{sec:frame}. Therefore, to optimize computational resource usage, we only used the current model framework for the subsequent departments (i.e. ophthalmology and TCM).

Fig.~\ref{fig:compare_training} depicts the efficacy of our \zalm on a session-wise inference example from the department of TCM. Since we have discussed the difficulty of objective assessment in Sec.~\ref{sec:objective}, it is challenging to match the model's responses directly with those of a real doctor. We thus mainly compare the model's outputs without and with the proposed \zalm, using the actual patient-doctor dialogue as a reference. As described in Sec.~\ref{sec:imple}, all text and image messages between the patient and the real doctor before the user's current message are regarded as the history input during the model training and inference. The original model, which was without \zalm, had the issues of premature diagnoses (``folliculitis'') and unnecessary referrals to other departments (``see a dermatologist''). In particular, when the patient sent a poor-quality tongue image, which was not a close-up view, the model incorrectly identified the affected area as the ``forehead''. In contrast, the model with \zalm generated acceptable responses to keep the conversation going and accurately addressed the appearance of the patient's tongue based on the cropped image, which was generally consistent with the real doctor's response. In this example, \zalm extracts the RoI region from the unsatisfactory image, enables the model's attention to shift to the tongue, and makes the responses more accurate.

\subsubsection{Medical VLM Frameworks}
\label{sec:frame}

In this section, we compare the performance of the initial and current VLM frameworks we developed. Since our assessment metric is a relative one, we record the mean opinion scores (MOS) values before deriving the DMOS. This absolute value can reveal the individual performance of the two VLM frameworks. We calculate the session-level averaged MOS$^{\text{ref}}$ and MOS, representing the original model without \zalm (i.e. the reference model) and the model with \zalm, respectively. Table~\ref{tab:mos} displays the performance results of these two frameworks. First, the reference result of Qwen-VL + Qwen-14B-Chat significantly outperforms that of BLIP2 + Baichuan2-13B-Chat (3.23 vs. 2.82), and is even also comparable to the result of BLIP2 + Baichuan2-13B-Chat equipped with \zalm (3.23 vs. 3.25). Moreover, the performance gap between the frameworks widens further after applying \zalm to Qwen-VL + Qwen-14B-Chat (3.25 vs. 3.71). Therefore we adopt Qwen-VL + Qwen-14B-Chat as the current VLM framework.

\begin{table}[t]
    \centering
    \caption{Performance of VLMs without \zalm (i.e. averaged MOS$^{\text{ref}}$) and with \zalm (i.e. averaged MOS) in dermatology. 
    \vspace{-1ex}
    }
    \begin{tabular}{cc}
    	\toprule
		VLM & Avg. MOS$^{\text{ref}}$$\uparrow$ / Avg. MOS$\uparrow$ \\ \midrule
        BLIP-2\cite{li2023blip} + Baichuan2-13B-Chat\cite{baichuan2023baichuan2} & 2.82 / \textbf{3.25} (out of 4) \\
        Qwen-VL\cite{bai2023qwen} + Qwen-14B-Chat\cite{qwen} &  3.23 / \textbf{3.71} (out of 4) \\
        \bottomrule
    \end{tabular}    
    \vspace{-3ex}
    \label{tab:mos}
\end{table}

\section{Discussion, Limitations, and Future Work}
\label{sec:disc}

\subsection{Clinical Relevance}
Since the unforeseen COVID-19 outbreak, online doctor consultation applications have gradually become an integral component of clinical practice and continue to play an essential role in the post-pandemic era \cite{yang2024zhongjing,chen2020exploring}. In developing countries, where healthcare resources are severely imbalanced, online consultations effectively enable patients in underdeveloped areas to access high-quality medical expertise. As the volume of online consultations grows, introducing AI-assisted manners becomes increasingly necessary to alleviate the overhead for human doctors. Recent research has primarily focused on text-only patient-doctor conversation scenarios. Unlike those studies, our scenario encompasses multimodal multi-turn medical consultations, incorporating both text and image modalities, which is a notable challenge in online doctor consultation applications that has yet to be well addressed. If responses to images cannot be generated effectively, it could severely harm the patient’s medical experience. For example in Fig.~\ref{fig:compare_training}, the red content generated by the ``original model'' would significantly downgrade the patient's satisfaction.

Our proposed \zalm has demonstrated its capacity to accurately identify and extract information from user-uploaded images, enhancing the quality of vision-language alignment during VLM training and inference. By enabling AI doctors to better understand patient descriptions of affected areas (through both text and images) and to more effectively interact with patients, \zalm can significantly improve diagnostic efficiency and accuracy in real clinical settings, further reducing the labor costs for human doctors.

\subsection{Privacy Constraints}
Many medical LLMs conducted performance comparisons with ChatGPT/GPT-4 or used them for scoring \cite{moor2023med,yang2024zhongjing,chen2023huatuogpt,singhal2023towards}. Unfortunately, due to our platform's medical data privacy protection policies, we are prohibited from calling external API interfaces including ChatGPT/GPT-4. However, those studies have already demonstrated that open-source LLMs trained with specialized medical data outperform general-purpose LLMs in professional fields. Given the stringent data privacy requirements in medical environments, \zalm's compliance with data protection policies ensures it can safely be integrated into clinical practice without compromising sensitive patient information.

\subsection{Model Selection in \zalm}
To achieve a better performance of keyword extractions in \zalm, we believe a model fine-tuned with medical data can be adopted, such as the LLM part in our department-wise VLM. It is worth noting that the images sent by users can be regarded as ``natural images with medical information'' rather than typical medical images (DICOM grayscale format). While there are some outstanding medical multimodal representation models \cite{huang2021gloria,wang2022multi,cheng2023prior}, a zero-shot visual grounding model trained on natural images, like GDINO, fits our scenario more.

\subsection{Limitations}
While our experiments across multiple departments have proven the efficacy of the proposed \zalm, there are also some limitations. As mentioned in Sec.~\ref{sec:imple}, due to computational limitations, we adopt a two-step strategy for our model training. With sufficient computational resources, we believe that fine-tuning our entire VLM with LoRA \cite{hu2021lora} on the database curated by \zalm will achieve better results \cite{zhu2023melo,wu2023next}. If sufficient computational resources are available, whether attaching a visual grounding model affects the performance of the VLM itself is an ongoing research area \cite{jiao2024enhancing}. Since our platform only includes Chinese patients and does not involve people of other skin tones or ethnicities, this may affect the universality of our proposed method to these populations. Due to the limitations of our practical application scenarios, we have conducted experiments on multimodal data in Chinese only. However, our method can be applied to multi-turn multimodal medical dialogue in other languages as well.

\subsection{Future Work}
We observe that some patients upload photos of medications and lab reports, which contain extensive textual information, requiring the model to have the capacity of optical character recognition (OCR). With the recent emergence of studies on medical document question answering (DQA) \cite{jin2024rjua}, addressing this issue becomes important in our future work. 

\section{Conclusions}
\label{sec:concl}

In this paper, we propose \zalm, a zero-shot scheme to improve vision-language alignment in the multi-turn multimodal medical dialogue scenario, rather than altering any VLM structure. Aiming to address the issue of poor-quality images sent by patients, \zalm leverages the preceding text conversations before an image that has semantic coherence to infer the RoIs in the image. It adopts an LLM to summarize the keywords from the prior texts and a visual grounding model to crop the RoIs, in order to enrich vision-language alignment. Compared with the commonly used but simplistic win-loss-tie rule, we design a new evaluation metric to convey a fine-grained performance comparison for subjective assessment of multi-turn unimodal/multimodal medical dialogue. 
Our experiments using two versions of LLMs on the in-house datasets from three different clinical departments statistically demonstrate the significant effectiveness of \zalm.

\section*{References}
\bibliographystyle{IEEEtran}
\bibliography{refer}

\newpage
\appendices

\section{Prompt for Keyword Extraction}
The prompt for keyword extraction in \zalm is provided in Table~\ref{tab:prompt}.

\begin{table}[ht]
	\centering
	\caption{Prompt for the LLM in \zalm to extract keywords from the preceding context of an image. 
        }
	\label{tab:prompt}
	\small
	\begin{tcolorbox}
		
		\textbf{Prompt:}\vspace{0.2cm}

		\begin{CJK}{UTF8}{gbsn}{你是一个在各科室临床医疗均有丰富经验的医疗助手，接下来患者将会发送若干段他们和医生交流时的对话，这些对话内容围绕着他们发送的病灶图像，你的任务是从患者的对话中提取出最多5个关键词，这些关键词将用于对病灶图像进行定位。回复我之前请确认只回复提取的英文关键词，不要描述对话，也不要回复任何其他内容。对话如下：}\vspace{0.1cm}\end{CJK}
		
		(\textit{Translation:} You are an experienced medical assistant with rich clinical experience in various departments. You will be sent several conversations between patients and doctors, which focus on the lesions images they send. Your task is to extract a maximum of 5 keywords from the patient's dialogue, which will be used to locate the lesion image. Please translate the keywords and only reply with the extracted keywords IN ENGLISH, not the conversation content, and do not reply to any other content. The conversation is as follows.)
		
		\vspace{0.1cm}
		[Input]\vspace{0.1cm}
		
		\begin{CJK}{UTF8}{gbsn}{“我怀孕后腿上起了这样的红块”，“特别是大腿内侧”，“左边不严重，这个是右边的”，“这是今天洗澡完拍的”}\vspace{0.1cm}\end{CJK}
		
		(\textit{Translation:} ``I got these red patches on my legs after I became pregnant'', ``Especially the inner thighs'', ``The left side is not serious, this is the right side'', ``This was taken after a bath today'')
		\vspace{0.1cm}
		[Output]\vspace{0.1cm}
		
		``Pregnancy'', ``legs'', ``red patches'', ``thighs'',  ``right side''

		\begin{CJK}{UTF8}{gbsn}(\textit{Translation:} {``怀孕''，``腿''，``红斑''，``大腿''，``右侧''})\end{CJK}

	\end{tcolorbox}
\end{table}

\section{Grounding DINO (GDINO)}

\begin{figure*}[ht]
    \centering
    \includegraphics[width=\linewidth]{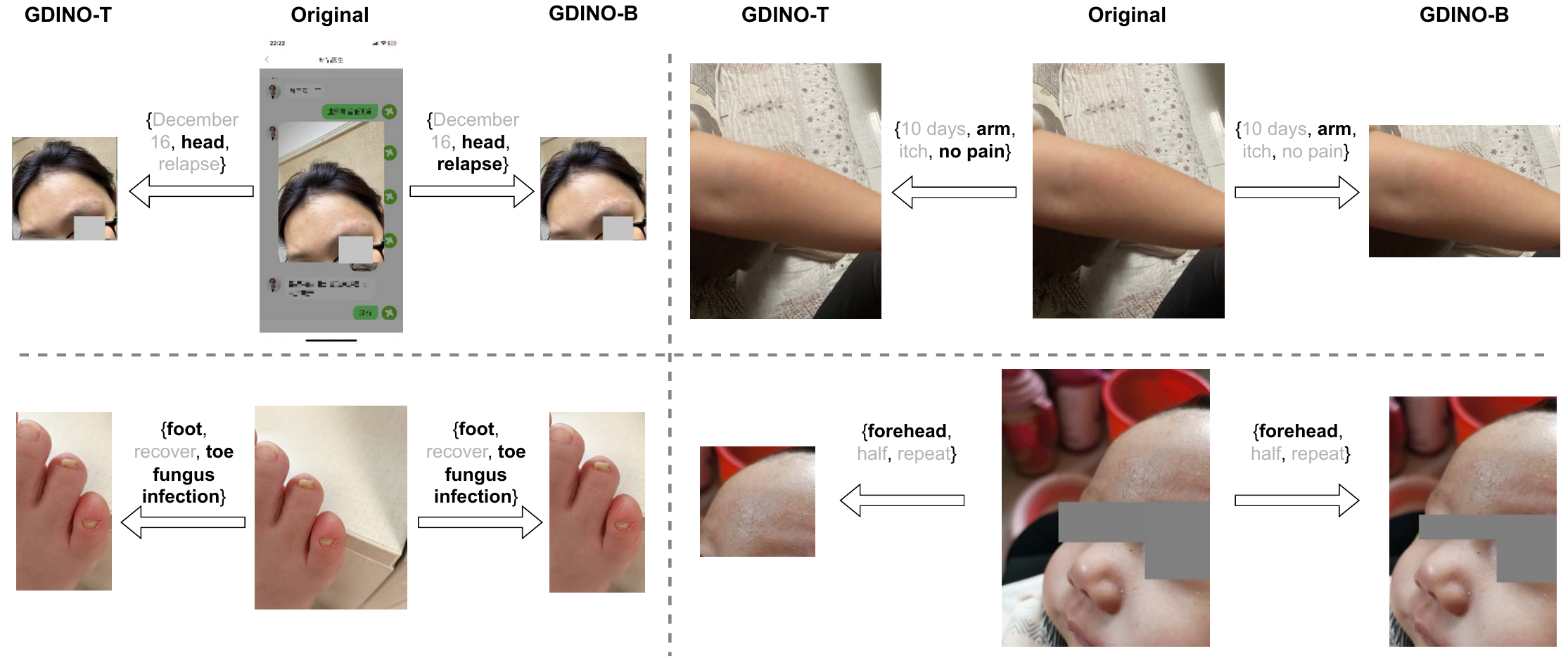}
    \caption{Illustrated examples from the department of dermatology using GDINO-T and GDINO-B for visual grounding operations using in-context information. The words or phrases with gray color are not activated by GDINO-T or GDINO-B. To protect patients' privacy, the gray patches on patients' eyes are manually added separately when we create this figure, rather than being generated by GDINO models.}
	\label{fig:gdino-crop}
\end{figure*}

Although GDINO has shown the performance of their Tiny, Base, and Large models \cite{liu2023grounding}, the available checkpoints are for the Tiny (GDINO-T) and Base (GDINO-B) models only.
Fig.~\ref{fig:gdino-crop} provides several examples from the department of dermatology to illustrate the visual grounding performance of GDINO-T and GDINO-B. Both GDINO-T and GDINO-B can ignore irrelevant keywords or phrases, such as ``December 16'' and ``10 days''. While the two examples show the equality of Compared with GDINO-T, GDINO-B shows its superior performance. 
The two examples on the left are relatively simple, as the background interference is minimal, allowing both GDINO-T and GDINO-B to capture a consistent RoI. However, in the example in the upper right corner, the extensive texture patterns in the background interfere with GDINO-T's region of interest (RoI) extraction. In the lower right example, GDINO-T tends to be more strict in extracting the RoIs, while GDINO-B is more lenient. Since we aim to remove irrelevant elements from the image without losing core information, we believe that GDINO-B performs better overall. This observation has also been endorsed by doctors.

\end{document}